\documentclass{llncs}

\usepackage{amsfonts}
\usepackage{amsmath}
\usepackage{arydshln}  
\usepackage[table]{xcolor}  
\usepackage{graphics}  
\usepackage{multirow}  
\def\b{\ensuremath\boldsymbol}
\usepackage{graphicx}

\usepackage{amssymb}

\usepackage{url}
\usepackage[algo2e,linesnumbered,lined,boxed,vlined]{algorithm2e}

\usepackage{tikz}
\usepackage{lipsum}
\newcommand\copyrighttextt{%
  \footnotesize Accepted (to appear) in International Conference on Image Analysis and Recognition (ICIAR) 2020, Springer.}
\newcommand\copyrightnotice{%
\begin{tikzpicture}[remember picture,overlay]
\node[anchor=south,yshift=10pt] at (current page.south) {\fbox{\parbox{\dimexpr\textwidth-\fboxsep-\fboxrule\relax}{\copyrighttextt}}};
\end{tikzpicture}%
}

\begin{document}

\title{Backprojection for Training Feedforward  Neural Networks in the Input and Feature Spaces}

\author{Benyamin Ghojogh,
Fakhri Karray,
Mark Crowley
}
\authorrunning{F. Author et al.}
%
\institute{Department of Electrical and Computer Engineering, \\ University of Waterloo, Waterloo, ON, Canada  \\
\email{\{bghojogh, karray, mcrowley\}@uwaterloo.ca} 
}



\maketitle              

\begin{abstract}
After the tremendous development of neural networks trained by backpropagation, it is a good time to develop other algorithms for training neural networks to gain more insights into networks. In this paper, we propose a new algorithm for training feedforward neural networks which is fairly faster than backpropagation. This method is based on projection and reconstruction where, at every layer, the projected data and reconstructed labels are forced to be similar and the weights are tuned accordingly layer by layer. The proposed algorithm can be used for both input and feature spaces, named as backprojection and kernel backprojection, respectively. This algorithm gives an insight to networks with a projection-based perspective. The experiments on synthetic datasets show the effectiveness of the proposed method.  
\end{abstract}

\copyrightnotice

\section{Introduction}

In one of his recent seminars, Geoffrey Hinton mentioned that after all of the developments of neural networks \cite{fausett1994fundamentals} and deep learning \cite{goodfellow2016deep}, perhaps it is time to move on from backpropagation \cite{rumelhart1986learning} to newer algorithms for training neural networks. Especially, now that we know why shallow \cite{soltanolkotabi2018theoretical} and deep \cite{allen2019learning} networks work very well and why local optima are fairly good in networks \cite{feizi2017porcupine}, other training algorithms can help improve the insights into neural nets. 
Different training methods have been proposed for neural networks, some of which are backpropagation \cite{rumelhart1986learning}, genetic algorithms \cite{montana1989training,leung2003tuning}, and belief propagation as in restricted Boltzmann machines \cite{hinton2006reducing}. 

A neural network can be viewed from a manifold learning perspective \cite{hauser2017principles}. Most of the spectral manifold learning methods can be reduced to kernel principal component analysis \cite{ham2004kernel} which is a projection-based method \cite{ghojogh2019unsupervised}. 
Moreover, at its initialization, every layer of a network can be seen as a random projection \cite{karimi2018exploring}. Hence, a promising direction could be a projection view of training neural networks. 
In this paper, we propose a new training algorithm for feedforward neural networks based on projection and \textit{backprojection} (or so-called reconstruction). 
In the backprojection algorithm, we update the weights layer by layer. For updating a layer $m$, we project the data from the input, until the layer $m$. We also backproject the labels of data from the last layer to the layer $m$. The projected data and backprojected labels at layer $m$ should be equal because in a perfectly trained network, projection of data by the entire layers should result in the corresponding labels. Thus, minimizing a loss function over the projected data and backprojected labels would correctly tune the layer's weights. 
This algorithm is proposed for both the input and feature spaces where in the latter, the kernel of data is fed to the network.

\section{Backprojection Algorithm}

\subsection{Projection and Backprojection in Network}

In a neural network, every layer without its activation function acts as a linear projection. Without the nonlinear activation functions, a network/autoencoder is reduced to a linear projection/principal component analysis \cite{ghojogh2019unsupervised}. 
If $\b{U}$ denotes the projection matrix (i.e., the weight matrix of a layer), $\b{U}^\top \b{x}$ projects $\b{x}$ onto the column space of $\b{U}$. 
The reverse operation of projection is called reconstruction or backprojection and is formulated as $\b{U} \b{U}^\top \b{x}$ which shows the projected data in the input space dimensionality (note that it is $\b{U}\b{f}^{-1}(\b{f}(\b{U}^\top \b{x}))$ if we have a nonlinear function $\b{f}(.)$ after the linear projection). 
At the initialization, a layer acts as a random projection \cite{karimi2018exploring} which is a promising feature extractor according to the Johnson-Lindenstrauss lemma \cite{achlioptas2003database}. Fine tuning the weights using labels makes the features more useful for discrimination of classes. 

\subsection{Definitions}

Let us have a training set $\mathcal{X} := \{\b{x}_i \in \mathbb{R}^d \}_{i=1}^n$ and their one-hot encoded labels $\mathcal{Y} := \{\b{y}_i \in \mathbb{R}^p \}_{i=1}^n$ where 
$n$, $d$, and $p$ are the sample size, dimensionality of data, and dimensionality of labels, respectively. 
We denote the dimensionality or the number of neurons in layer $m$ by $d_m$. By convention, we have $d_0 := d$ and $d_{n_\ell} = p$ where $n_\ell$ is the number of layers and $p$ is the dimensionality of the output layer. Let the data after the activation function of the $m$-th layer be denoted by $\b{x}^{(m)} \in \mathbb{R}^{d_m}$. 
Let the projected data in the $m$-th layer be $\mathbb{R}^{d_m} \ni \b{z}^{(m)} := \b{U}_m^\top\, \b{x}^{(m-1)}$ where $\b{U}_m \in \mathbb{R}^{d_{m-1} \times d_m}$ is the weight matrix of the $m$-th layer. 
Note that $\b{x}^{(m)} = \b{f}_m(\b{z}^{(m)})$ where $\b{f}_m(.)$ is the activation function in the $m$-th layer. By convention, $\b{x}^{(0)} := \b{x}$.
The data are projected and passed through the activation functions layer by layer; hence, $\b{x}^{(m)}$ is calculated as:
\begin{align}
\mathbb{R}^{d_m} \ni \b{x}^{(m)} := \b{f}_{m}( \b{U}_m^\top\, \b{f}_{m-1}(\b{U}_{m-1}^\top\, \cdots \b{f}_{1}(\b{U}_{1}^\top \b{x}) ) )  = \b{f}_{m}( \b{U}_m^\top\, \b{x}^{(m-1)} ).
\end{align}
In a mini-batch gradient descent set-up, let $\{\b{x}_i\}_{i=1}^b$ be a batch of size $b$. 
For a batch, we denote the outputs of activation functions at the $m$-th layer by $\mathbb{R}^{d_m \times b} \ni \b{X}^{(m)} := [\b{x}_1^{(m)}, \dots, \b{x}_b^{(m)}]$.

Now, consider the one-hot encoded labels of batch, denoted by $\b{y} \in \mathbb{R}^p$. 
We take the inverse activation function of the labels and then reconstruct or \textit{backproject} them to the previous layer to obtain $\b{y}^{(n_\ell-1)}$. We do similarly until the layer $m$. Let $\b{y}^{(m)} \in \mathbb{R}^{d_m}$ denote the backprojected data at the $m$-th layer, calculated as:
\begin{align}
\b{y}^{(m)}\! :=\! \b{U}_{m+1}\, \b{f}_{m+1}^{-1}( \b{U}_{m+2}\, \b{f}_{m+2}^{-1}( \cdots \b{U}_{n_\ell}\, \b{f}_{n_\ell}^{-1}(\b{y}) )  )\! =\! \b{U}_{m+1}\, \b{f}_{m+1}^{-1}(\b{y}^{(m+1)}).
\end{align}
By convention, $\b{y}^{(n_\ell)} := \b{y}$. The backprojected batch at the $m$-th layer is $\mathbb{R}^{d_m \times b} \ni \b{Y}^{(m)} := [\b{y}_1^{(m)}, \dots, \b{y}_b^{(m)}]$. 
We use $\b{X} \in \mathbb{R}^{d \times b}$ and $\b{Y} \in \mathbb{R}^{p \times b}$ to denote the column-wise batch matrix and its one-hot encoded labels. 

\subsection{Optimization}

In the backprojection algorithm, we optimize the layers' weights one by one. Consider the $m$-th layer whose loss we denote by $\mathcal{L}_m$:
\begin{align}\label{equation_backprojection_loss}
\underset{\b{U}_m}{\text{minimize}} \quad \mathcal{L}_m := \sum_{i=1}^b \ell(\b{x}_i^{(m)} - \b{y}_i^{(m)}) = \sum_{i=1}^b \ell \big(\b{f}_m(\b{U}_m^\top\, \b{x}_i^{(m-1)}) - \b{y}_i^{(m)}\big),
\end{align}
where $\ell(.)$ is a loss function such as the squared $\ell_2$ norm (or Mean Squared Error (MSE)), cross-entropy, etc. The loss $\mathcal{L}_m$ tries to make the projected data $\b{x}_i^{(m)}$ as similar as possible to the backprojected data $\b{y}_i^{(m)}$ by tuning the weights $\b{U}_m$. 
This is because the output of the network is supposed to be equal to the labels, i.e., $\b{x}^{(n_\ell)} \approx \b{y}$. 
In order to tune the weights for Eq. (\ref{equation_backprojection_loss}), we use a step of gradient descent. 
Using chain rule, the gradient is:
\begin{align}
\mathbb{R}^{d_{m-1} \times d_m} \ni \frac{\partial \mathcal{L}_m}{\partial \b{U}_m} =\! \sum_{i=1}^b \textbf{vec}^{-1}_{d_{m-1} \times d_m}\! \Big[ \big(\frac{\partial \b{z}_i^{(m)}}{\partial \b{U}_m}\big)^\top \big(\frac{\partial \b{f}_m(\b{z}_i^{(m)})}{\partial \b{z}_i^{(m)}}\big)^\top \frac{\partial \ell(\b{f}_m(\b{z}_i^{(m)}))}{\partial \b{f}_m(\b{z}_i^{(m)})} \Big],
\end{align}
where we use the Magnus-Neudecker convention in which matrices are vectorized and $\textbf{vec}^{-1}_{d_{m-1} \times d_m}$ is de-vectorization to $d_{m-1} \times d_m$ matrix.
If the loss function is MSE or cross-entropy for example, the derivatives of the loss function w.r.t. the activation function, respectively, are:
\begin{align}
&\mathbb{R}^{d_m} \ni \frac{\partial \ell(\b{f}_m(\b{z}_i^{(m)}))}{\partial \b{f}_m(\b{z}_i^{(m)})} = 2 \big(\b{f}_m(\b{z}_i^{(m)}) -  \b{y}_i^{(m)}\big), \text{ and} \\
&\mathbb{R}^{d_m} \ni \frac{\partial \ell(\b{f}_m(\b{z}_i^{(m)}))}{\partial \b{f}_m(\b{z}_i^{(m)})} = -\Big[\frac{\b{y}_{i,j}^{(m)}}{\b{f}_m(\b{z}_{i,
j}^{(m)})}, \forall j \in \{1, \dots, d_m\}\Big]^\top,
\end{align}
where $\b{y}_{i,j}^{(m)}$ and $\b{z}^{(m)}_{i,j}$ are the $j$-th dimension of $\b{y}_{i}^{(m)}$ and $\b{z}^{(m)}_{i} = \b{U}_m^\top\, \b{x}_i^{(m-1)}$, respectively. 

For the activation functions in which the nodes are independent, such as linear, sigmoid, and hyperbolic tangent, the derivative of the activation function w.r.t. its input is a diagonal matrix:
\begin{align}
\mathbb{R}^{d_m \times d_m} \ni \frac{\partial \b{f}_m(\b{z}_i^{(m)})}{\partial \b{z}_i^{(m)}} = 
\textbf{diag}\Big(\frac{\partial \b{f}_{m}(\b{z}^{(m)}_{i,j})}{\partial \b{z}^{(m)}_{i,j}}, \forall j \in \{1, \dots, d_m\}\Big),
\end{align}
where $\textbf{diag}(.)$ makes a matrix with its input as diagonal. 

The derivative of the projected data before the activation function (i.e., the input of the activation function) w.r.t. the weights of the layer is:
\begin{align}
\mathbb{R}^{d_m \times (d_m d_{m-1})} \ni \frac{\partial \b{z}_i^{(m)}}{\partial \b{U}_m} = \frac{\partial\, \b{U}_m^\top\, \b{x}_i^{(m-1)}}{\partial \b{U}_m} = \b{I}_{d_m} \otimes \b{x}_i^{(m-1)\top},
\end{align}
where $\otimes$ denotes the Kronecker product and $\b{I}_{d_m}$ is the $d_m \times d_m$ identity matrix. 

The procedure for updating weights in the $m$-the layer is shown in Algorithm \ref{algorithm_update_layer}. 
Until the layer $m$, data is projected and passed through activation functions layer by layer. Also, the label is backprojected and passed through inverse activation functions until the layer $m$. A step of gradient descent is used to update the layer's weights where $\eta > 0$ is the learning rate. 
Note that the backprojected label at a layer may not be in the feasible domain of its inverse activation function. Hence, at every layer, we should project the backprojected label onto the feasible domain \cite{parikh2014proximal}. We denote projection onto the feasible set by $\Pi(.)$.

\SetAlCapSkip{0.5em}
\IncMargin{0.8em}
\begin{algorithm2e}[!t]
\DontPrintSemicolon
    \textbf{Procedure: } UpdateLayerWeights($\mathcal{U}$, $\b{X}$, $\b{Y}$, $m$)\;
    \textbf{Input: } weights: $\mathcal{U} := \{\b{U}_r\}_{r=1}^{n_\ell}$, batch data: $\b{X} \in \mathbb{R}^{d \times b}$, batch labels: $\b{Y} \in \mathbb{R}^{p \times b}$, layer: $m \in [1, n_\ell]$\;
    $\b{X}^{(0)} := \b{X}$\;
    \For{layer $r$ from $1$ to $(m-1)$}{
        $\b{Z}^{(r)} := \b{U}_r^\top \b{X}^{(r-1)}$\;
        $\b{X}^{(r)} := \b{f}_r( \b{Z}^{(r)} )$\;
    }
    $\b{Y}^{(n_\ell)} := \b{Y}$\;
    \For{layer $r$ from $(n_\ell-1)$ to $m$}{
        $\b{Y}^{(r+1)} := \Pi(\b{Y}^{(r+1)})$\;
        $\b{Y}^{(r)} := \b{U}_{r+1}\, \b{f}_{r+1}^{-1}(\b{Y}^{(r+1)})$\;
    }
    $\b{U}_m := \b{U}_m - \eta\, (\partial \mathcal{L}_m / \partial \b{U}_m)$\;
    \textbf{Return} $\b{U}_m$\;
\caption{Updating the weights of a layer in backprojection}\label{algorithm_update_layer}
\end{algorithm2e}
\DecMargin{0.8em}

\SetAlCapSkip{0.5em}
\IncMargin{0.8em}
\begin{algorithm2e}[!h]
\DontPrintSemicolon
    \textbf{Procedure: } Backprojection($\mathcal{X}$, $\mathcal{Y}$, $b$, $e$)\;
    \textbf{Input: } training data: $\mathcal{X}$, training labels: $\mathcal{Y}$, batch size: $b$, number of epochs: $e$ \;
    Initialize $\mathcal{U} = \{\b{U}_r\}_{r=1}^{n_\ell}$\;
    \For{epoch from $1$ to $e$}{
        \For{batch from $1$ to $\lceil n / b \rceil$}{
            $\b{X}, \b{Y} \leftarrow$ take batch from  $\mathcal{X}$ and $\mathcal{Y}$\;
            \uIf{procedure is forward}{
                \For{layer $m$ from $1$ to $n_\ell$}{
                    $\b{U}_m \leftarrow$ UpdateLayerWeights($\{\b{U}_r\}_{r=1}^{n_\ell}$, $\b{X}$, $\b{Y}$, $m$)\;
                }
            }
            \uElseIf{procedure is backward}{
                \For{layer $m$ from $n_\ell$ to $1$}{
                    $\b{U}_m \leftarrow$ UpdateLayerWeights($\{\b{U}_r\}_{r=1}^{n_\ell}$, $\b{X}$, $\b{Y}$, $m$)\;
                }
            }
            \ElseIf{procedure is forward-backward}{
                \uIf{batch index is odd}{
                    \For{layer $m$ from $1$ to $n_\ell$}{
                        $\b{U}_m \leftarrow$ UpdateLayerWeights($\{\b{U}_r\}_{r=1}^{n_\ell}$, $\b{X}$, $\b{Y}$, $m$)\;
                    }
                }
                \Else{
                    \For{layer $m$ from $n_\ell$ to $1$}{
                        $\b{U}_m \leftarrow$ UpdateLayerWeights($\{\b{U}_r\}_{r=1}^{n_\ell}$, $\b{X}$, $\b{Y}$, $m$)\;
                    }
                }
            }
        }
    }
\caption{Backprojection}\label{algorithm_backprojection}
\end{algorithm2e}
\DecMargin{0.8em}

\subsection{Different Procedures}

So far, we explained how to update the weights of a layer. Here, we detail updating the entire network layers.
In terms of the order of updating layers, we can have three different procedures for a backprojection algorithm. One possible procedure is to update the first layer first and move to next layers one by one until we reach the last layer. Repeating this procedure for the batches results in the \textit{forward procedure}. 
In an opposite direction, we can have the \textit{backward procedure} where, for each batch, we update the layers from the last layer to the first layer one by one. 
If we have both directions of updating, i.e., forward update for a batch and backward update for the next batch, we call it the \textit{forward-backward procedure}. 
Algorithm \ref{algorithm_backprojection} shows how to update the layers in different procedures of the backprojection algorithm. 
Note that in this algorithm, an updated layer impacts the update of next/previous layer. One alternative approach is to make updating of layers dependent only on the weights tuned by previous mini-batch. 
In that approach, the training of layers can be parallelized within mini-batch.

\section{Kernel Backprojection Algorithm}

Suppose $\b{\phi}: \mathcal{X} \rightarrow \mathcal{H}$ is the pulling function to the feature space. Let $t$ denote the dimensionality of the feature space, i.e., $\b{\phi}(\b{x}) \in \mathbb{R}^t$. 
Let the matrix-form of $\mathcal{X}$ and $\mathcal{Y}$ be denoted by $\mathbb{R}^{d \times n} \ni \breve{\b{X}} := [\b{x}_1, \dots, \b{x}_n]$ and $\mathbb{R}^{p \times n} \ni \breve{\b{Y}} := [\b{y}_1, \dots, \b{y}_n]$. 
The kernel matrix \cite{hofmann2008kernel} for the training data $\breve{\b{X}}$ is defined as $\mathbb{R}^{n \times n} \ni \breve{\b{K}} := \b{\Phi}(\breve{\b{X}})^\top \b{\Phi}(\breve{\b{X}})$ where $\mathbb{R}^{t \times n} \ni \b{\Phi}(\breve{\b{X}}) := [\b{\phi}(\b{x}_1), \dots, \b{\phi}(\b{x}_n)]$. 
We normalize the kernel matrix \cite{ah2010normalized} as $\breve{\b{K}}(i,j) := \breve{\b{K}}(i,j) / \big[\breve{\b{K}}(i,i) \breve{\b{K}}(j,j)\big]^{1/2}$ where $\breve{\b{K}}(i,j)$ denotes the $(i,j)$-th element of the kernel matrix.

According to representation theory \cite{alperin1993local}, the projection matrix $\b{U}_1 \in \mathbb{R}^{d \times d_1}$ can be expressed as a linear combination of the projected training data. Hence, we have $\mathbb{R}^{t \times d_1} \ni \b{\Phi}(\b{U}_1) = \b{\Phi}(\breve{\b{X}})\, \b{\Theta}$ where every column of $\b{\Theta} := [\b{\theta}_1, \dots, \b{\theta}_{d_1}] \in \mathbb{R}^{n \times d_1}$ is the vector of coefficients for expressing a projection direction as a linear combination of projected training data. 
The projection of the pulled data is $\mathbb{R}^{d_1 \times n} \ni \b{\Phi}(\b{U}_1)^\top \b{\Phi}(\breve{\b{X}}) = \b{\Theta}^\top \b{\Phi}(\breve{\b{X}})^\top \b{\Phi}(\breve{\b{X}}) = \b{\Theta}^\top \breve{\b{K}}$.

In the kernel backprojection algorithm, in the first network layer, we project the pulled data from the feature space with dimensionality $t$ to another feature space with dimensionality $d_1$. The projections of the next layers are the same as in backprojection. In other words, \textit{kernel backprojection applies backprojection in the feature space rather than the input space.} 
In a mini-batch set-up, we use the columns of the normalized kernel corresponding to the batch samples, denoted by $\{\b{k}_i \in \mathbb{R}^{n}\}_{i=1}^b$. Therefore, the projection of the $i$-th data point in the batch is $\mathbb{R}^{d_1} \ni \b{\Theta}^\top \b{k}_i$. In kernel backprojection, the dimensionality of the input is $n$ and the kernel vector $\b{k}_i$ is fed to the network as input. 
If we replace the $\b{x}_i$ by $\b{k}_i$, Algorithms \ref{algorithm_update_layer} and \ref{algorithm_backprojection} are applicable for kernel backprojection. 

In the test phase, we normalize the kernel over the matrix $[\breve{\b{X}}, \b{x}_t]$ where $\b{x}_t \in \mathbb{R}^{d}$ is the test data point. Then, we take the portion of normalized kernel which correspond to the kernel over the training versus test data, denoted by $\mathbb{R}^{n} \ni \b{k}_t := \b{\Phi}(\breve{\b{X}})^\top \b{\Phi}(\b{x}_t)$. The projection at the first layer is then $\mathbb{R}^{d_1} \ni \b{\Theta}^\top \b{k}_t$. 

\begin{figure}[!t]
  \makebox[\textwidth][c]{\includegraphics[width=1.5\textwidth]{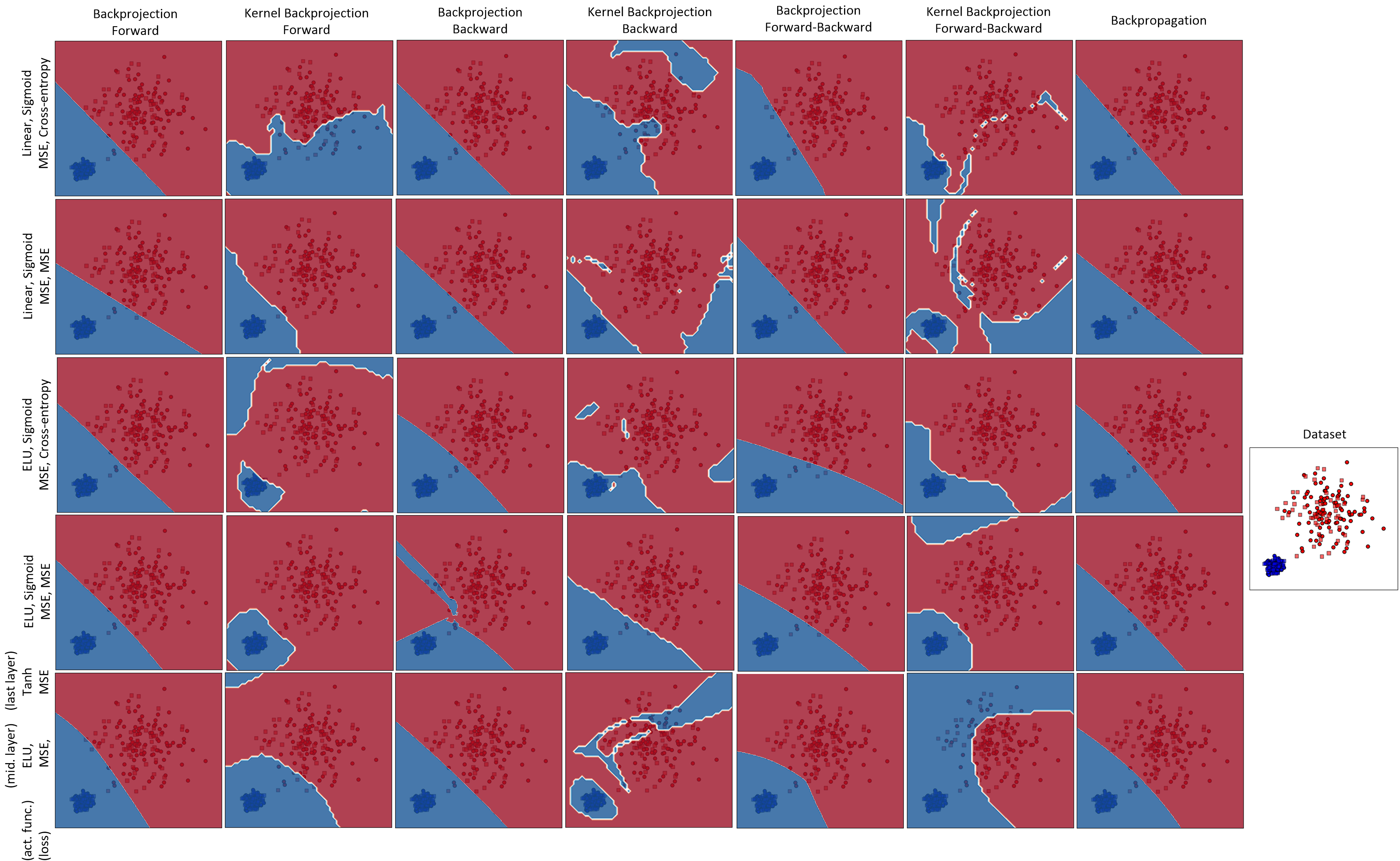}}%
  \caption{Discrimination of two classes by different training algorithms with various activation functions and loss functions. The label for each row indicates the activation functions and the loss functions for the middle then the last layers.}
  \label{figure_dataset_twoClass}
\end{figure}

\section{Experiments}

\begin{figure}[!t]
  \makebox[\textwidth][c]{\includegraphics[width=1.4\textwidth]{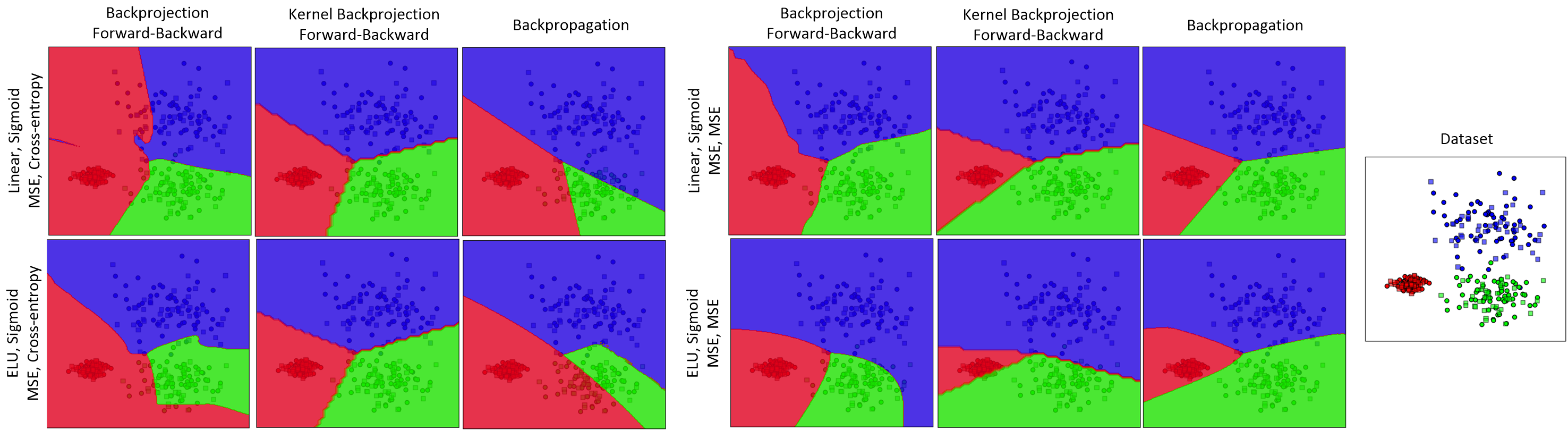}}%
  \caption{Discrimination of three classes by different training algorithms with various activation functions and loss functions.}
  \label{figure_dataset_threeClass}
\end{figure}

\noindent
\textbf{Datasets:}
For experiments, we created two synthetic datasets with 300 data points each, one for binary-class and one for three-class classification (see Figs. \ref{figure_dataset_twoClass} and \ref{figure_dataset_threeClass}). 
For more difficulty, we set different variances for the classes. 
The data were standardized as a preprocessing. 
For this conference short-paper, we limit ourselves to introduction of this new approach and small synthetic experiments. Validation on larger real-world datasets is ongoing for future publication. 

\hfill \break
\noindent
\textbf{Neural Network Settings:}
We implemented a neural network with three layers whose number of neurons are $\{15, 20, p\}$ where $p=1$ and $p=3$ for the binary and ternary classification, respectively. 
In different experiments, we used MSE loss for the middle layers and MSE or cross-entropy losses for the last layer. 
Moreover, we used Exponential Linear Unit (ELU) \cite{clevert2015fast} or linear functions for activation functions of the middle layers while sigmoid or hyperbolic tangent (tanh) were used for the last layer. 
The derivative and inverse of these activation functions are as the following:
\begin{align*}
& \text{ELU: } f(z) = 
\left\{
    \begin{array}{ll}
        e^z - 1, z \leq 0 \\
        z, z > 0
    \end{array}
\right.,
f'(z) = 
\left\{
    \begin{array}{ll}
        e^z, z \leq 0 \\
        1, z > 0
    \end{array}
\right.,
f^{-1}(y) = 
\left\{
    \begin{array}{ll}
        \ln(y+1), y \leq 0 \\
        y, y > 0
    \end{array}
\right., \\
& \text{Linear: } f(z) = z, \quad f'(z) = 1, \quad f^{-1}(y) = y, \\
& \text{Sigmoid: } f(z) = \frac{1}{1 + e^{-z}}, \quad f'(z) = f(1-f), \quad f^{-1}(y) = \ln(\frac{y}{1-y}), \\
& \text{Tanh: } f(z) = \frac{e^z - e^{-z}}{e^{z} + e^{-z}}, \quad f'(z) = 1 - f^2, \quad f^{-1}(y) = 0.5 \ln(\frac{1+y}{1-y}),
\end{align*}
where in the inverse functions, we bound the output values for computational reasons in computer. 
Mostly, a learning rate of $\eta=10^{-4}$ was used for backprojection and backpropagation and $\eta=10^{-5}$ was used for kernel backprojection.

\noindent
\textbf{Comparison of Procedures:}
The performance of different forward, backward, and forward-backward procedures in backprojection and kernel backprojection are illustrated in Fig. \ref{figure_dataset_twoClass}. 
In these experiments, the Radial Basis Function (RBF) kernel was used in kernel backprojection. 
Although the performance of these procedures are not identical but all of them are promising discrimination of classes. 
This shows that all three proposed procedures work well for backprojection in the input and feature spaces. In other words, the algorithm is fairly robust to the order of updating layers.

\hfill \break
\noindent
\textbf{Comparison to Backpropagation:}
The performances of backprojection, kernel backprojection, and backpropagation are compared in the binary and ternary classification, shown in Figs. \ref{figure_dataset_twoClass} and \ref{figure_dataset_threeClass}, respectively. 
In Fig. \ref{figure_dataset_threeClass}, the linear kernel was used. 
In Fig. \ref{figure_dataset_twoClass}, most often, kernel backprojection considers a spherical class around the blue (or even red) class which is because of the choice of RBF kernel. 
Comparison to backpropagation in the two figures shows that backprojection's performance nearly matches that of backpropagation. 

In the different experiments, the mean time of every epoch was often 0.08, 0.11, and 0.2 seconds for backprojection, kernel backprojection, and backpropagation, respectively, where the number of epochs were fairly similar in the experiments. This shows that backprojection is \textit{faster} than backpropagation. This is because backpropagation updates the weights one by one while backprojection updates layer by layer.  




\section{Conclusion and Future Direction}\label{section_conclusion}

In this paper, we proposed a new training algorithm for feedforward neural network named backprojection.
The proposed algorithm, which can be used for both the input and feature spaces, tries to force the projected data to be similar to the backprojected labels by tuning the weights layer by layer. 
This training algorithm, which is moderately faster than backpropagation in our initial experiments, can be used with either forward, backward, or forward-backward procedures. 
It is noteworthy that adding a penalty term for weight decay \cite{krogh1992simple} to Eq. (\ref{equation_backprojection_loss}) can regularize the weights in backprojection \cite{ghojogh2019theory}. Moreover, batch normalization can be used in backprojection by standardizing the batch at the layers \cite{ioffe2015batch}. 
This paper concentrated on feedforward neural networks. As a future direction, we can develop backprojection for other network structures such as convolutional networks \cite{lecun1998gradient} and carry more expensive validation experiments on real-world data. 

\bibliographystyle{splncs}      
\bibliography{references.bib}            

\begin{thebibliography}{1}

\bibitem{sample-citation}
Last, F., Last2, F.:
\newblock Canadian {{AI}}.
\newblock Canadian Artificial Intelligence \textbf{31} (2018)  1--12

\end{thebibliography}


\begin{thebibliography}{10}

\bibitem{fausett1994fundamentals}
Fausett, L.:
\newblock Fundamentals of neural networks: architectures, algorithms, and
  applications.
\newblock Prentice-Hall, Inc. (1994)

\bibitem{goodfellow2016deep}
Goodfellow, I., Bengio, Y., Courville, A.:
\newblock Deep learning.
\newblock MIT press (2016)

\bibitem{rumelhart1986learning}
Rumelhart, D.E., Hinton, G.E., Williams, R.J.:
\newblock Learning representations by back-propagating errors.
\newblock Nature \textbf{323}(6088) (1986)  533--536

\bibitem{soltanolkotabi2018theoretical}
Soltanolkotabi, M., Javanmard, A., Lee, J.D.:
\newblock Theoretical insights into the optimization landscape of
  over-parameterized shallow neural networks.
\newblock IEEE Transactions on Information Theory \textbf{65}(2) (2018)
  742--769

\bibitem{allen2019learning}
Allen-Zhu, Z., Li, Y., Liang, Y.:
\newblock Learning and generalization in overparameterized neural networks,
  going beyond two layers.
\newblock In: Advances in neural information processing systems. (2019)
  6155--6166

\bibitem{feizi2017porcupine}
Feizi, S., Javadi, H., Zhang, J., Tse, D.:
\newblock Porcupine neural networks: (almost) all local optima are global.
\newblock arXiv preprint arXiv:1710.02196 (2017)

\bibitem{montana1989training}
Montana, D.J., Davis, L.:
\newblock Training feedforward neural networks using genetic algorithms.
\newblock In: IJCAI. Volume~89. (1989)  762--767

\bibitem{leung2003tuning}
Leung, F.H.F., Lam, H.K., Ling, S.H., Tam, P.K.S.:
\newblock Tuning of the structure and parameters of a neural network using an
  improved genetic algorithm.
\newblock IEEE Transactions on Neural networks \textbf{14}(1) (2003)  79--88

\bibitem{hinton2006reducing}
Hinton, G.E., Salakhutdinov, R.R.:
\newblock Reducing the dimensionality of data with neural networks.
\newblock Science \textbf{313}(5786) (2006)  504--507

\bibitem{hauser2017principles}
Hauser, M., Ray, A.:
\newblock Principles of {Riemannian} geometry in neural networks.
\newblock In: Advances in neural information processing systems. (2017)
  2807--2816

\bibitem{ham2004kernel}
Ham, J.H., Lee, D.D., Mika, S., Sch{\"o}lkopf, B.:
\newblock A kernel view of the dimensionality reduction of manifolds.
\newblock In: International Conference on Machine Learning. (2004)

\bibitem{ghojogh2019unsupervised}
Ghojogh, B., Crowley, M.:
\newblock Unsupervised and supervised principal component analysis: Tutorial.
\newblock arXiv preprint arXiv:1906.03148 (2019)

\bibitem{karimi2018exploring}
Karimi, A.H.:
\newblock Exploring new forms of random projections for prediction and
  dimensionality reduction in big-data regimes.
\newblock Master's thesis, University of Waterloo (2018)

\bibitem{achlioptas2003database}
Achlioptas, D.:
\newblock Database-friendly random projections: {Johnson}-{Lindenstrauss} with
  binary coins.
\newblock Journal of computer and System Sciences \textbf{66}(4) (2003)
  671--687

\bibitem{parikh2014proximal}
Parikh, N., Boyd, S.:
\newblock Proximal algorithms.
\newblock Foundations and Trends{\textregistered} in Optimization \textbf{1}(3)
  (2014)  127--239

\bibitem{hofmann2008kernel}
Hofmann, T., Sch{\"o}lkopf, B., Smola, A.J.:
\newblock Kernel methods in machine learning.
\newblock The annals of statistics (2008)  1171--1220

\bibitem{ah2010normalized}
Ah-Pine, J.:
\newblock Normalized kernels as similarity indices.
\newblock In: Pacific-Asia Conference on Knowledge Discovery and Data Mining,
  Springer (2010)  362--373

\bibitem{alperin1993local}
Alperin, J.L.:
\newblock Local representation theory: Modular representations as an
  introduction to the local representation theory of finite groups. Volume~11.
\newblock Cambridge University Press (1993)

\bibitem{clevert2015fast}
Clevert, D.A., Unterthiner, T., Hochreiter, S.:
\newblock Fast and accurate deep network learning by exponential linear units
  ({ELUs}).
\newblock In: International Conference on Learning Representations (ICLR).
  (2016)

\bibitem{krogh1992simple}
Krogh, A., Hertz, J.A.:
\newblock A simple weight decay can improve generalization.
\newblock In: Advances in neural information processing systems. (1992)
  950--957

\bibitem{ghojogh2019theory}
Ghojogh, B., Crowley, M.:
\newblock The theory behind overfitting, cross validation, regularization,
  bagging, and boosting: tutorial.
\newblock arXiv preprint arXiv:1905.12787 (2019)

\bibitem{ioffe2015batch}
Ioffe, S., Szegedy, C.:
\newblock Batch normalization: Accelerating deep network training by reducing
  internal covariate shift.
\newblock arXiv preprint arXiv:1502.03167 (2015)

\bibitem{lecun1998gradient}
LeCun, Y., Bottou, L., Bengio, Y., Haffner, P.:
\newblock Gradient-based learning applied to document recognition.
\newblock Proceedings of the IEEE \textbf{86}(11) (1998)  2278--2324

\end{thebibliography}

\end{document}